\documentclass[numbered]{trbunofficial}
\usepackage{graphicx}
\usepackage{algorithm}
\usepackage{algpseudocode}
\usepackage{xcolor}
\usepackage[hidelinks]{hyperref}
\usepackage{amsmath}
\usepackage{amsfonts}
\usepackage{tabularx} 
\usepackage{adjustbox} 
\usepackage{booktabs} 
\usepackage{array}    
\usepackage{longtable}

\AuthorHeaders{J.R. Palit et al 2025}
\title{Data-Driven Analysis of Crash Patterns in SAE Level 2 and Level 4 Automated Vehicles Using K-means Clustering and Association Rule Mining}

\author{%
\textbf{Jewel Rana Palit}, PE, RSP1(Corresponding Author)\\
Traffic Engineer/Project Manager-II\\
Collier County Government\\
Traffic Management Center\\
2695 Francis Ave Unit D, Naples, Fl, 37221\\
Email: jewel.palit@collier.gov\\
\hfill\break%
\textbf{Vijayalakshmi K Kumarasamy, Ph.D.}\\
Department of Computer Science and Engineering\\ 
University of Tennessee at Chattanooga\\
Chattanooga, TN, USA 37403\\
Email: lry466@mocs.utc.edu\\
\hfill\break%
\textbf{Osama A. Osman}, Ph.D., PE\\
ORCiD: https://orcid.org/0000-0002-5157-2805\\
Chief Scientist\\
Center of Urban Informatics and Progress\\
Chattanooga, TN, USA 37403\\
Email: osama-osman@utc.edu
}

\begin{document}
\maketitle
\section{Abstract}
Automated Vehicles (AV) hold potential to reduce or eliminate human driving errors, enhance traffic safety, and support sustainable mobility. Recently, crash data has increasingly revealed that AV behavior can deviate from expected safety outcomes, raising concerns about the technology’s safety and operational reliability in mixed traffic environments. While past research has investigated AV crash, most studies rely on small-size California-centered datasets, with a limited focus on understanding crash trends across various SAE Levels of automation. This study analyzes over 2,500 AV crash records from the United States National Highway Traffic Safety Administration (NHTSA), covering SAE Levels 2 and 4, to uncover underlying crash dynamics.A two-stage data mining framework is developed.  K-means clustering is first applied to segment crash records into 4 distinct behavioral clusters based on temporal, spatial, and environmental factors. Then, Association Rule Mining (ARM) is used to extract interpretable multivariate relationships between crash patterns and crash contributors including lighting conditions, surface condition, vehicle dynamics, and environmental conditions within each cluster. These insights provide actionable guidance for AV developers, safety regulators, and policymakers in formulating AV deployment strategies and minimizing crash risks.

\hfill\break%
\noindent\textit{Keywords}: Automated Vehicles, SAE Level, Crash Analysis, K-means Clustering, ADS, ADAS, Association Rule Mining. 
\newpage

\section{Introduction}
Over the last decade, traffic accidents have claimed 3,50,000 lives in the USA, with 40,990 fatalities in 2023 alone \cite{USDOT_2024_new}. Traffic accidents often result from human errors such as speeding, fatigue, and inattention, which are major contributors to fatal crashes. According to the National Highway Traffic Safety Administration (NHTSA), adopting automated vehicle (AV) technology could substantially reduce human errors, thereby decreasing crash frequency \cite{NHTSA_2025_new}. Additionally, AVs can play a key role in creating more mobile and environmentally sustainable traffic flow. With safety in mind, vehicles are currently equipped with different automation systems to assist drivers. These include Advanced Driver Assistance Systems (ADAS) and Automated Driving Systems (ADS) \cite{NHTSA_2025_new}. ADAS encompasses features such as Adaptive Cruise Control (ACC), Blind Spot Detection, and Lane Departure Warning, which support the driver during vehicle operation. In contrast, ADS involves higher levels of automation, allowing the system to manage driving without human intervention under specified conditions. The Society of Automotive Engineers (SAE) and NHTSA categorize these advanced technologies into six levels of automation, ranging from Level 0 (no automation) to Level 5 (full automation). ADAS corresponds to Levels 1 and 2, while ADS represents automation Levels 3, 4, and 5 \cite{NHTSA_2025_new, Int_S_Taxonomy_2025_new}. Despite the promise of enhanced safety and advanced assistance options available, AVs are not immune to traffic crashes, particularly when interacting with heterogeneous road users in mixed traffic settings. Even under ideal conditions, AVs have been involved in crashes, raising concerns and posing significant challenges to AV commercialization \cite{lee2020black, lee2024typical}. Researchers have observed that AV crashes do not follow typical traffic crash patterns, underscoring the importance of thoroughly understanding these patterns to identify limitations and improve the technology \cite{lee2024typical}.\\

To reveal the hidden patterns in these AV crashes, scholars have employed various data mining, statistical analysis, clustering, text mining techniques, etc. In 2015, Schoettle et al. conducted a preliminary statistical analysis of crash reports from self-driving vehicles operated by Google, Delphi, and Audi in California. They found that AVs had a higher crash rate per million miles compared to conventional vehicles. This study's findings are constrained by an extremely small sample size ($N=11$) \cite{schoettle2015preliminary}. Favaro et al. used traditional statistical analysis to investigate the California Department of Motor Vehicle (CA DMV) dataset. Their study indicated that rear-end collisions are the most frequent collision type. However, this study was also subjected to a limited sample size ($N=26$) and a single geographic location \cite{favaro2017examining}. Moreover, the limitations include reliance on predefined fields and manual review bias, as well as missing data for critical factors such as weather, lighting, and road conditions in crash narratives before their formal inclusion in the database in 2018 \cite{alambeigi2020crash}. To overcome the limitations of basic statistical methods in effectively understanding the crash pattern, Alambeigi et al. investigated the California DMV crash database using a Probabilistic Topic Modeling (PTM) approach. They identified five key crash scenarios from 114 AV crashes, with the most prevalent being rear-end collisions at intersections and driver-initiated transition crashes, highlighting critical safety concerns in AV operation and human-machine interactions \cite{alambeigi2020crash}. Though topic modeling provides insights from crash narratives, it is often prone to context sensitivity and offers a more generalized view of themes, potentially overlooking detailed and specific relationships. To deal with this shortcoming, supervised machine learning methods and Bayesian analysis, in addition to statistical and text analytics methods, have been used by several scholars \cite{ren2021divergent, boggs2020exploratory, kutela2022mining}. Kutela et al. used the supervised machine learning classifiers and Text Network Analysis (TNA) to investigate vulnerable road user (VRU) involvement in 252 AV crashes of the CA DMV, revealing crashes having positive relationships with crosswalks at intersections, signals, etc. \cite{kutela2022mining}. However, using Supervised Machine Learning (ML) methods for such a small-scale dataset could not facilitate extensive analysis for exploring hidden patterns of these crashes. Lee et al. used a Probabilistic Bayesian approach and TNA to investigate AV crash patterns and showed that AVs are more vulnerable to rear-end collisions in autonomous driving mode than in conventional mode \cite{lee2023advancing}. They also proved that manual disengagement tends to occur more often when an autonomous vehicle engages with a transit vehicle shortly before a collision. However, the Probabilistic Bayesian model used in the study is computationally intensive. Probabilistic Bayesian approach also requires the specification of prior distributions or modeling assumptions, which can introduce subjectivity and potential bias \cite{dunson2009bayesian}.\\

Several studies used clustering methods to ignore subjectivity in capturing AV crash patterns. Das et al. applied clustering analysis to understand AV crash patterns considering turning movement, manner of collisions, etc. \cite{das2020automated}. They explored six different collision patterns by analyzing 151 California DMV crash reports through Bayesian Clustering approaches \cite{das2020automated}. However, such clustering approaches struggle with capturing complex and evolving patterns over time \cite{kang2020gratis}. Clustering reveals co-occurring item groupings but does not explain their interrelationships. To understand how different factors in AV crashes affect each other, recent studies in AV have adopted Association Rule Mining (ARM), a data mining method that identifies interesting relations between different variables, either alone or in conjunction with clustering \cite{kumbhare2014overview}. By analyzing the co-occurrence of different items in the dataset, ARM can identify intricate relations, patterns, and associations among a set of items in short or large databases \cite{kumbhare2014overview}. ARM-generated rules reveal hidden trends and relationships between crash contributing factors, helping to develop strategies for reducing crash risk \cite{qin2025cracking, ashraf2021extracting}. Li et al. mapped 615 California DMV crashes and leveraged ARM to investigate pre-crash scenarios \cite{li2025characteristics}. Qin et al. used ARM to analyze nearly 1193 similar crash reports and found that AV crashes occur frequently at intersections with higher heterogeneous road user interactions \cite{qin2025cracking}. In 2021, Ashraf et al. combined ARM methods with Decision Tree model to extract the pre-crash rules of 251 CA DMV AV-involved crashes and analyze patterns of AV and non-AV movement in mixed traffic conditions \cite{ashraf2021extracting}. Lee et al. used ARM with hierarchical clustering to derive different accident scenarios from 165 California DMV crash reports \cite{lee2024typical}. Table \ref{tab:avcrash_analysis_summary} summarizes AV crash analysis approaches over the years. 
\subsection{Research Gap \& Contribution}
Significant efforts have been made by scholars over the years to investigate intrinsic automated vehicle crash patterns and crash factor relationships. However, most of the studies used basic statistical approaches. While combining ARM and clustering methods provided useful information, most of them used small-scale datasets, which are not sufficient and result in a fragmented knowledge domain \cite{ding2024exploratory}. Most studies are geographically skewed, as they use only California DMV-centered data, limiting the ability to reveal inherent crash patterns of automated vehicles on a national scale. Also, very few current studies draw a clear picture of crash patterns for different levels of automation equipped with advanced support systems (ADS \& ADAS) \cite{ding2024exploratory}. To capture ADS and ADAS crash patterns, Ding et al. first leveraged 1,001 SAE Level 2 (with ADAS) and 548 SAE Level 4 (with ADS) crash reports from the NHTSA database \cite{ding2024exploratory}. They proposed a Multinomial Logit model focusing only on injury severity as the target variable. Although this method captured linear relationships of AV crash factors and injury severity, it overlooked latent multivariate interactions of contributing crash factors.Yan et al. clustered 1374 ADS and ADAS-involved NHTSA crashes, emphasizing the co-occurrence of contributing factors rather than exploring their interrelationships \cite{yan2024comparison}. Understanding the multifaceted trend of automated vehicle crashes requires investigation of nonlinear dependencies across crash variables such as crash type, AV model, weather, temporal factors, vehicle dynamics, etc. It is crucial to move beyond co-occurrence analysis and focus on the relational dynamics among these variables.\\

To address this gap, we apply Association Rule Mining (ARM), a rule-based data mining technique, in conjunction with an unsupervised Machine Learning algorithm, K-means Clustering, to elucidate multivariate relationships within large-scale AV crash data. We utilize more than 2,500 SAE Level 2 (with ADAS) and Level 4 (with ADS) crash records from NHTSA to develop a more refined, nationally representative nuance of automated vehicle crash dynamics. We extract critical temporal variables through feature engineering, enabling the identification of high-confidence and context-aware associations among crash factors. Our hybrid framework clusters homogenous subgroups of crash contributors and leverages ARM to provide robust, context-specific, and explainable insights, supporting regulatory decision-making, policymakers, and AV developers.

\section{METHODOLOGY}
\begin{figure}[!ht]
  \centering
  \includegraphics[width=0.9\textwidth, height=0.32\textheight]{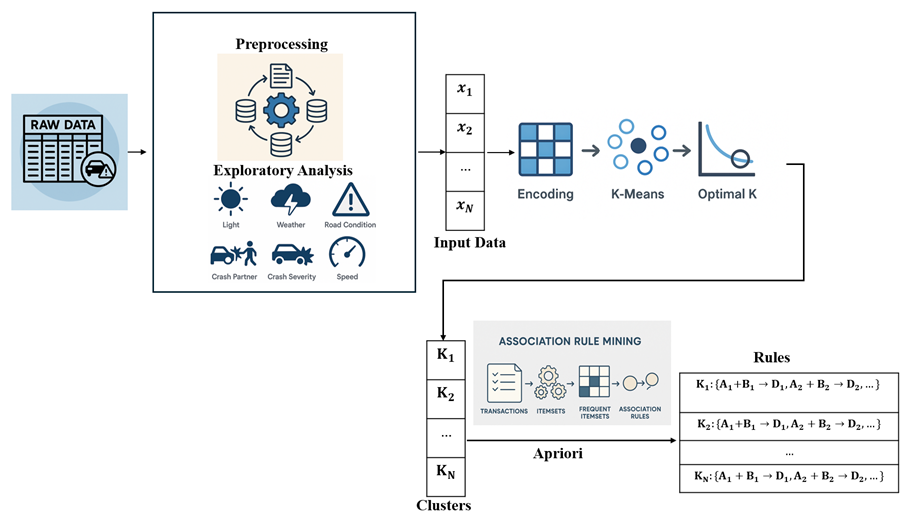}
  \caption{AV Crash Analysis Architecture}
  \label{fig:avcrash_analysis_arch}
\end{figure}

This study provides a data-driven pipeline to investigate multivariate associations among AV-related crashes equipped with ADAS and ADS support systems. We process the raw data and cluster it through the K-means algorithm. Then ARM is applied on separate clusters to generate rules enabling a better understanding of hidden AV crash patterns for different automation levels. The architecture of the methodology is briefly described in Figure\ref{fig:avcrash_analysis_arch}.

\begin{table}[H]
\caption{Summary of AV Crash Analysis Approaches from Literature Review}
\label{tab:avcrash_analysis_summary}
\begin{adjustbox}{width=\textwidth}
\begin{tabular}{|p{3.2cm}|p{2cm}|p{1.5cm}|p{3.2cm}|p{4cm}|p{4cm}|}
\hline
\textbf{Author} & \textbf{Data} & \textbf{Sample Size} & \textbf{Method} & \textbf{Helpful Aspect} & \textbf{Limitation} \\
\hline
Schoettle et al., 2015 & CA DMV & 11 & Basic Statistics & Comparison of the crash trends of AVs to conventional cars. & Data size is insufficient. Only one geographic location is considered. \\
\hline
Alambeigi et al., 2020 & CA DMV & 114 & Probabilistic Topic Modeling (PTM) & Identification of the most frequent crash using the crash narrative. & Only one geographic location is considered. PTM is subjected to Context sensitivity, leading to subjective bias. \\
\hline
Lee et al., 2023 & CA DMV & 260 & Probabilistic Bayesian Model & Exploration of pre-crash conditions and their relationship. & Bayesian models are prone to subjectivity of prior assumptions. Only one geographic location is considered. \\
\hline
Das et al., 2020 & CA DMV & 151 & Bayesian Clustering & Extension of the work of Favaro et al. with a Bayesian approach that categorized six distinct collision patterns. & Authors cluster crash factors together but do not explain how they are related. The study is geographically skewed. \\
\hline
Li et al., 2025 & CA DMV & 615 & ARM & Authors establish multivariate relationships of pre-crash factors. & The study is geographically skewed. Authors ignore different automation level characteristics with ADS \& ADAS. \\
\hline
Qin et al., 2025 & CA DMV & 1193 & ARM & Authors investigate the effect of heterogeneous road users in AV crashes. & The study is geographically skewed. Authors ignore different automation level characteristics with ADS \& ADAS. \\
\hline
Ashraf et al., 2021 & CA DMV & 251 & ARM \& Decision Tree & Authors combine ARM with ML algorithms to generate rules of pre-crash factors in mixed traffic conditions. & ML models are not effective for small data sizes. Only one geographic location is considered. Authors ignore different automation level characteristics with ADS \& ADAS. \\
\hline
Lee et al., 2024 & CA DMV & 165 & ARM \& Hierarchical Clustering & Scenario-based testing using a hybrid approach. & The study is geographically skewed. Authors ignore different automation level characteristics with ADS \& ADAS. \\
\hline
Ding et al., 2024 & NHTSA \& other sources & 1549 & Multinomial Logit & Nationwide AV crash data involvement. Analyze different automation levels of ADS \& ADAS. & Authors overlook multivariate associations of crash factors and only focus on injury severity. \\
\hline
\end{tabular}
\end{adjustbox}
\end{table}

\subsection{Data}
To enhance federal oversight, in 2021, NHTSA issued its first Standard General Order (SGO), which mandates that AV manufacturers and operators report crashes involving vehicles equipped with ADAS and ADS systems \cite{NHTSA_gov_2025}. This established a standardized and nationwide repository for AV crash data. We use this NHTSA data from 2021 through July 2024. Raw data includes Level 2 (with ADAS), Level 3 \& 4 (with ADS) automated vehicle crash records, totaling 2,703 rows and 137 different elements. To facilitate the data analysis process, ADS \& ADAS data are merged on the same columns, and numerous data variables having similar information have been compressed and unified, resulting in 2703 rows and 36 columns. A snapshot of the raw dataset is presented in Table \ref{tab:rawdata_nhtsa_sgo}, and Table \ref{tab:nhtsa_sgo_var_desc} provides brief descriptions of the variables.

\begin{table}[H]
\centering
\caption{Snapshot of Raw NHTSA SGO Data}
\label{tab:rawdata_nhtsa_sgo}
\begin{adjustbox}{width=\textwidth}
\begin{tabular}{|l|l|l|l|l|l|l|l|l|l|}
\hline
\textbf{Report ID} & \textbf{Reporting Entity} & \textbf{Incident Time} & \textbf{State} & \textbf{ADS/ADAS Version} & \textbf{Make} & \textbf{Telematics} & \textbf{Media} & \textbf{Narrative} & \dots \\
\hline
30270 & Waymo & 20:35:00 & CA & 5th Gen & Jaguar & Y &  & On April X, 2024, a & \dots \\
\hline
28349 & Lucid & 03:38:00 & CA & Level 2 & Lucid &  & Y & May X, & \dots \\
\hline
30270 & Tesla, Inc. & 02:21:00 & TX & 0 & Tesla & Y &  & On January X, 2022 & \dots \\
\hline
\multicolumn{10}{|c|}{\dots} \\
\hline
\end{tabular}
\end{adjustbox}
\end{table}

\begin{longtable}{|p{4cm}|p{2.5cm}|p{9cm}|}
\caption{NHTSA SGO Variable Descriptions}
\label{tab:nhtsa_sgo_var_desc} \\
\hline
\textbf{Variable} & \textbf{Type} & \textbf{Description} \\
\hline
\endfirsthead

\multicolumn{3}{c}%
{{\bfseries Table \thetable{} -- continued from previous page}} \\
\hline
\textbf{Variable} & \textbf{Type} & \textbf{Description} \\
\hline
\endhead

\hline \multicolumn{3}{r}{{Continued on next page}} \\
\endfoot

\hline
\endlastfoot

Report ID & Text & A unique ID for each report issued by NHTSA. \\
Reporting Entity & Text & Name of the entity filing the report \\
State & Text & Name of the State the incident took place \\
City & Text & Name of the City the incident took place \\
Latitude & Number & Geographic Coordinate \\
Longitude & Number & Geographic Coordinate \\
Model year & Number & Model year of the subject AV involved in the crash. \\
Mileage & Number & Mileage of the subject AV involved in the crash. \\
Incident Time & Number & Time of the incident (24:00) occurrence reported by the reporting entity. \\
Automation System & Text & ADAS/ADS equipped in the subject vehicle at the time of the incident. \\
Weather & Text & Indicates the weather during the time and place of the incident (cloudy, clear, etc.). \\
Roadway Description & Text & The roadway conditions, excluding weather and surface conditions (work zone, missing marking, traffic incident, etc.) \\
Roadway Type & Text & Type of road where the incident took place (intersection, highway/freeway, etc.) \\
Roadway Surface & Text & Condition of road where the incident took place (dry, wet, unknown, etc.) \\
Lighting & Text & Lighting conditions during the time and at the location of the incident (daylight, dark-lighted, dawn/dusk, etc.) \\
Source & Text & Reporting Entity’s source via which it first received notice of the incident (telematics, complaint, law enforcement, etc.) \\
Highest Injury Severity Alleged & Text & Reporting Entity’s reported highest confirmed or alleged injury level (minor, fatality, no injuries reported, etc.) \\
CP Pre-Crash Movement & Text & Pre-crash movement pattern of any involved crash partner before the incident (making a right turn, left turn, stopped, etc.) \\
CP Contact Area & Text & Reported damaged area of the other vehicle or crash partner (CP) that encountered the subject vehicle (front, rear, left \& front left, etc.) \\
SV Contact Area & Text & Reported damaged area of the subject AV vehicle (front, rear, etc.) \\
SV Precrash Speed & Number & Reported speed of the subject vehicle (SV)/AV in mph before the incident. \\
Any Air Bags Deployed? & Text & Reporting Entity’s report of whether any air bag deployed during the incident in the crash partner / Other vehicle. \\
Posted Speed Limit & Number & Road speed limit in mph during the time and at the location of the incident. \\
Crash With & Text & Any vehicle, non-motorist, animal, or object with which the subject vehicle came into contact during the incident (Passenger Car, SUV, Van, Pickup, etc.) \\
\multicolumn{3}{|c|}{\dots} \\
\end{longtable}

\subsection{Data Preparation and Exploratory Analysis}
\begin{figure}[!ht]
  \centering
  \includegraphics[width=0.8\textwidth, height=0.7\textheight]{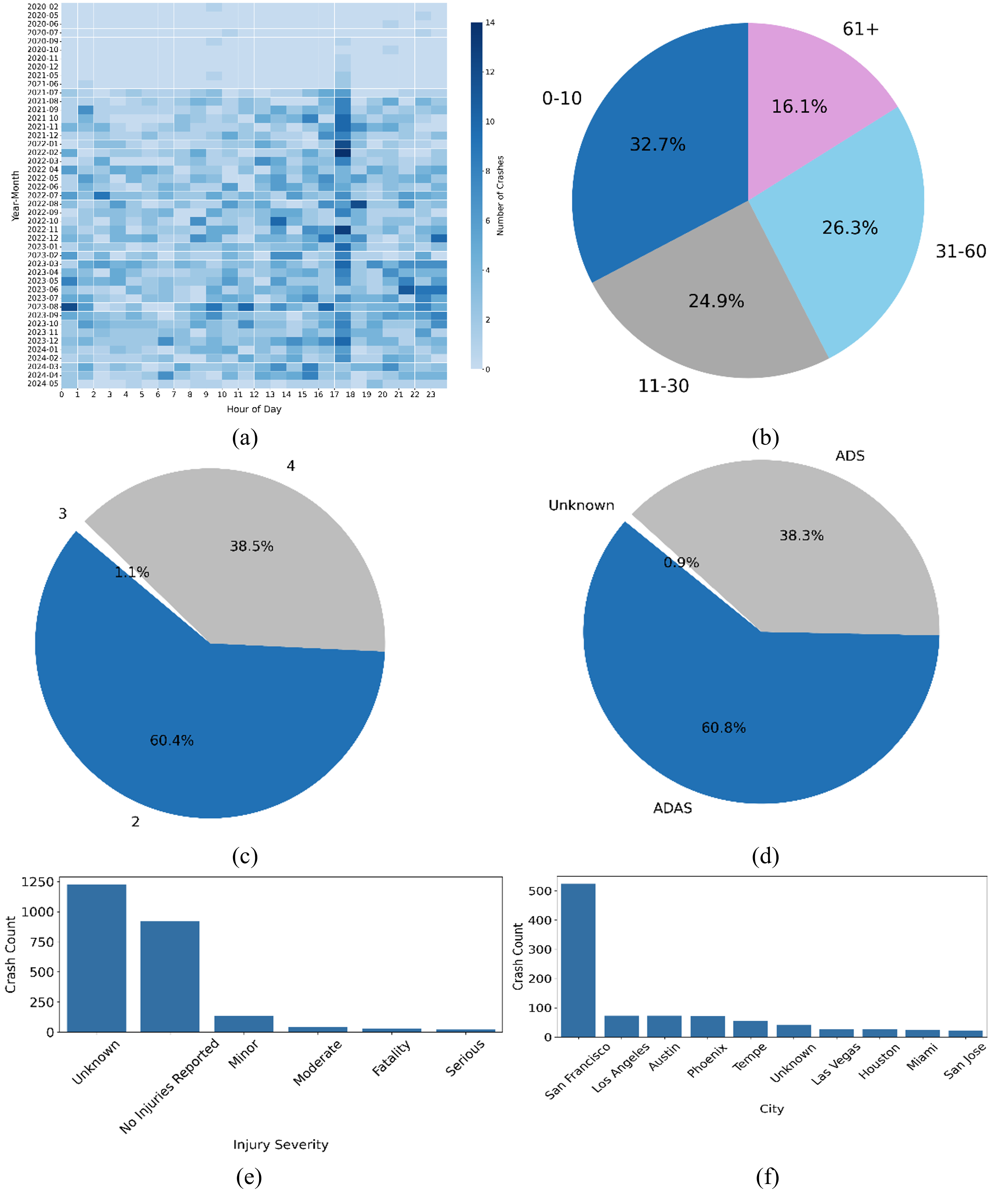}
  \caption{Exploratory Analysis (a) Heatmap of Incident Time, (b)SV Pre Crash Speed Range (mph), (c) Crash \% Distribution on Automation Level, (d) Crash \% Distribution on Automation System ADS/ADAS, (e) Highest Injury Severity Alleged, (f) AV Crashes for top 10 city}
  \label{fig:exploratory_analysis}
\end{figure}

We process the initially merged data (2703 rows and 36 columns) through missing value handling, redundant data reduction (investigating officer, address, VIN, etc.), and feature engineering. First, we filter out the duplicate entries based on the unique Report ID and then discard this variable from the filtered data, as it does not affect crashes. The simplest solution for handling missing values is to get rid of the data; however, it decreases the data size significantly \cite{kaiser2014dealing}. We get rid of the variables with significant missing values (more than 50\%) and impute others with the mode to retain information \cite{kim1977treatment} for numerical variables and put ‘Unknown’ for categorical variables. We applied a simplified text mining technique on crash narratives to capture information on the automation level (Level 2, 3 \& 4) and created a separate variable. We also broke down the incident date and time into variables such as year, month of the year, day of the month, and hour of the day to capture temporal relationships. Information in the variables like speed and mileage has been separated into different bins (e.g., 0-10 mph, <10 mph, 0-10 k, etc.) for the convenience of data encoding. The final refined dataset has a size of 2380×31 which includes mixed categorical and numerical variables such as SV Precrash Speed, Lighting, Weather, Highest Injury Severity Alleged, Hour, Day, Month, Year, Source, CP Contact Area, Automation System (e.g., ADS, ADAS), Automation Level (e.g., 2, 3, 4), etc. Figure \ref{fig:exploratory_analysis} shows some preliminary statistics of the cleaned data. From Figure \ref{fig:exploratory_analysis}(a), it is evident that most of the crashes occurred between 2 PM and 7 PM, showing a crash hotspot during evening rush hours while people mostly return from work. We also see another hotspot area between September 2021 and February 2023. Figure \ref{fig:exploratory_analysis}(b) shows that almost 50\% of crashes occurred when the AV was operating within 11-60 mph. Figure \ref{fig:exploratory_analysis}(c) \& (d) show the dominance of Level 2 (with ADAS) and SAE Level 4 (with ADS). The remaining levels of automation are minimal and therefore excluded from subsequent analysis. Figure \ref{fig:exploratory_analysis}(e) shows that most crashes are minor injuries and harmless. Figure \ref{fig:exploratory_analysis}(f) indicates most of the crashes are centered in SF and LA (CA-centered) followed by Austin \& Phoenix. \\

\begin{figure}[!ht]
  \centering
  \includegraphics[width=0.9\textwidth, height=0.32\textheight]{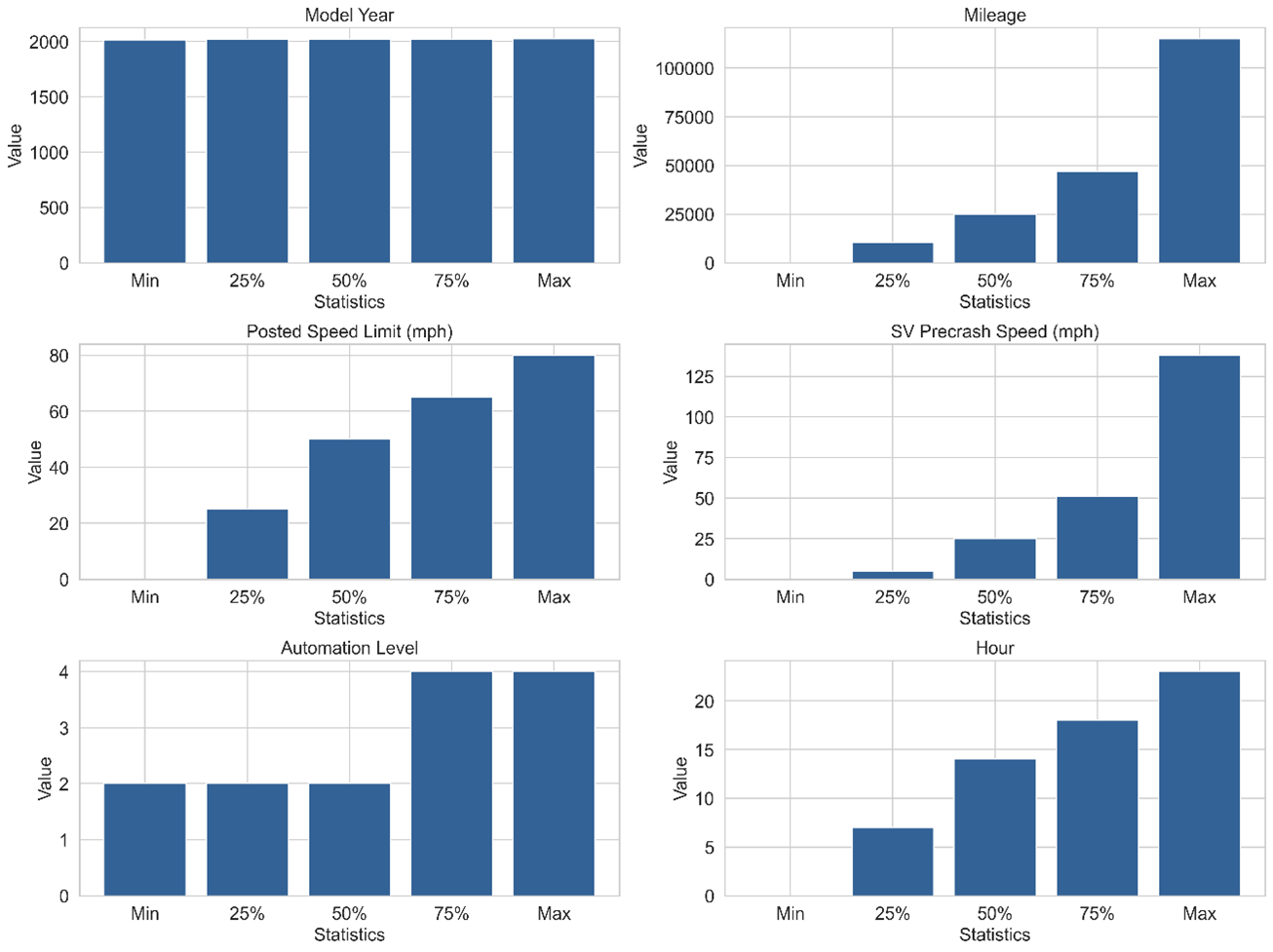}
  \caption{Descriptive Statistics of Numeric Variable}
  \label{fig:descriptive_Stat_numericvars}
\end{figure}

\begin{figure}[!ht]
  \centering
  \includegraphics[width=0.9\textwidth, height=0.32\textheight]{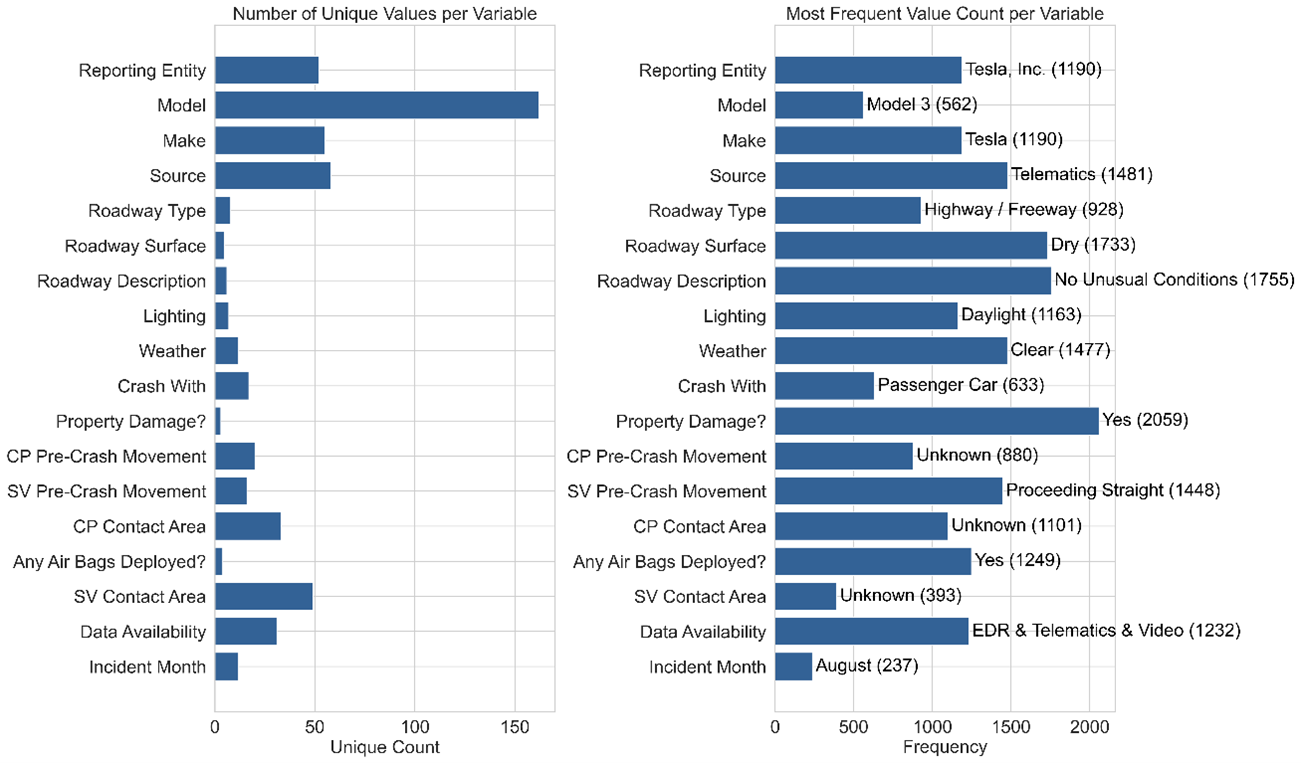}
  \caption{Descriptive Statistics of Categorical Variable}
  \label{fig:descriptive_Stat_categoricalvars}
\end{figure}

The descriptive statistics of some other numeric and categorical variables are shown in Figure \ref{fig:descriptive_Stat_numericvars} and Figure \ref{fig:descriptive_Stat_categoricalvars}. Figure \ref{fig:descriptive_Stat_numericvars} indicates the highest variance in mileage. Most AV crashes occur within a 65-mph speed limit. Figure \ref{fig:descriptive_Stat_categoricalvars} shows Tesla as the top reporting entity that is involved in the highest number of crashes. With 160 unique values, we see a wide range of operating AV models available on the market. Most of the incidents occurred during daylight, and the weather was clear. It also shows that before getting involved in crashes, the subject vehicle/AV was proceeding straight, and most of the time airbags were deployed. Most incidents seem to occur during August, which is the school starting season, representing the peak travel demand time.


\subsection{K-means Clustering}
To reveal hidden patterns of AV crashes and reduce intra-class heterogeneity in the subsequent ARM phase, we apply the unsupervised K-means clustering method. This allows us to segment the dataset into different subsets having shared crash behavior. We choose K-means as our ideal clustering method because the K-means algorithm is suitable for moderate to large datasets like ours, which provides efficient speed and scalability \cite{jain2010data}. While it is a simple approach, it offers an intuitive interpretation of the classified clusters and is efficient for high arrays of data where other complex ML models will be an overkill \cite{xu2005survey}. We apply label encoding (numeric type) for all categorical fields, as numerical approaches are suitable for distance-based ML models like K-means. The encoded dataset of size $2380 \times 31$ is fed into the K-means algorithm to partition it into K number of distinct clusters and minimize within-cluster variance. The algorithm starts by choosing K initial centroids with the following objective function, also known as within-cluster sum of squares (WCSS)(Eq. \ref{eq:kmeans_clustering}):
\begin{equation}
\label{eq:kmeans_clustering}
\min(Z_1, \ldots, Z_K) = \sum_{k=1}^{K} \sum_{x_i \in Z_k}^{x_n \in Z_k} \|x_i - y_k\|^2
\end{equation}

Here, $Z_k$ represents a set of points assigned to a cluster $k$, $x_i, \ldots, x_n$ are the data points inside $Z_k$ and $y_k$ is the centroid of $Z_k$. After choosing $K$ initial centroids, the algorithm calculates the Euclidean distance $\|x_i - y_k\|^2$ and assigns each point to the closest (minimum) centroid \cite{morissette2013k, burkardt2009k}. The algorithm keeps iterating this process until the centroids don’t change anymore and converge at $y_k^{(t)} = y_k^{(t-1)}$ for all $k$ where $t=$ iteration number. We calculate WCSS from $K = 2$ to $K = 10$ and determine the optimum $K$ value through the "elbow" method, which is the inflection point of the WCSS curve \cite{cui2020introduction}.

\subsection{Association Rule Mining (ARM)}
\begin{figure}[!ht]
  \centering
  \includegraphics[width=0.9\textwidth, height=0.2\textheight]{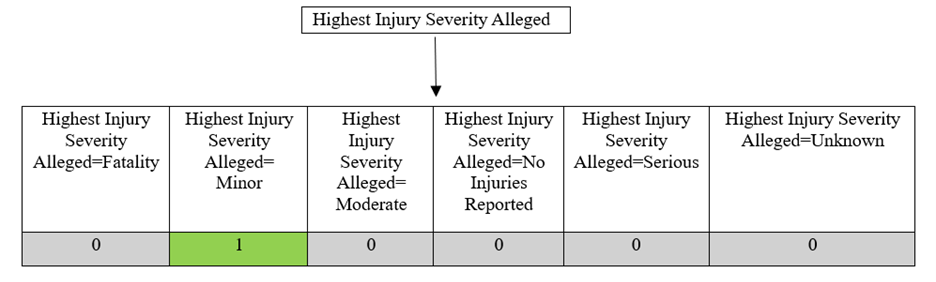}
  \caption{One Hot Encoded Data Size Conversion}
  \label{fig:onehot_encoded_data}
\end{figure}

After optimum cluster K is achieved, we apply ARM on each separate cluster to reveal interpretable multivariate associations between vehicle dynamics, weather conditions, and other crash contributors. Since ARM works on binary input, we first convert our data into a one-hot encoded format, where each category is transformed into a 0 or 1 indicator. For example, “Highest Injury Severity Alleged $=$ Minor” becomes a column with 1 if true and 0 otherwise. If we consider the variable “Highest Injury Severity Alleged,” which has 6 unique types, it is broken down in the following way where every row or crash record with respect to the “Highest Injury Severity Alleged” field having a $1 \times 1$ size will have a $1 \times 6$ size and can have a binary value of 1 and 0 as shown in Figure \ref{fig:onehot_encoded_data} with following probable possibilities.

We select the Apriori Algorithm for this study, as it is an efficient and robust ARM method suitable for moderate to large-scale datasets \cite{lee2024typical, hahsler2015mining}. Consider $T$ as a set of all transactions or a set of all crash records and
$I = \{i_1, i_2, i_3, \ldots, i_k\}$, a set of $k$ distinct items. Each transaction is associated with a single unique identifier. For any frequent itemset $X: X \subseteq I$, an association rule is represented by an antecedent and consequent, which generally denote a condition similar to "If" (left side of the condition or rule) and "Then" (right side of the condition or rule) respectively \cite{hahsler2019arulesviz, hahsler2015mining}. For each frequent itemset $X$, the generated rule is $A \rightarrow B$ where $A \in I$ and $B \in I$. Typically, 3 metrics are used for ARM rules: Support (S), Confidence (C) and Lift (L) \cite{hahsler2015mining}. Support is a measure of how frequently an itemset or association appears in the full dataset. Let us consider a simple association between $A$ and $B$ is such that $A \rightarrow B$. The support of this rule is formulated by Eq. \ref{eq:support_of_AimpliesB}:

\begin{equation}
\label{eq:support_of_AimpliesB}
S(A \rightarrow B) = \frac{\text{transactions having both } A \text{ and } B}{\text{total number of transaction, } N} = \frac{\text{count}(A \text{ and } B)}{N} = \frac{n(AB)}{N} = P(A \cap B)
\end{equation}\\

Confidence measures the probability of $B$ occurring when $A$ is present, reflecting the reliability of the association and formulated by Eq. \ref{eq:confidence_of_AimpliesB}:

\begin{equation}
\label{eq:confidence_of_AimpliesB}
C(A \rightarrow B) = \frac{S(A \cup B)}{S(A)} = P(B|A)
\end{equation}\\

Lift calculates the strength of the rule expressing how much the antecedent $A$ affects the occurrence of $B$, comparing the collective occurrence of $A$ and $B$ to the expected frequency if both of these were statistically independent. If $A$ and $B$ are fully independent, $L = 1$, expressing no effect of the antecedent. $L > 1$ shows $A$ increases the likelihood of $A$ and positive associations with each other, while $L < 1$ shows negative association. The formula is shown in Eq.\ref{eq:lift_of_AimpliesB}:

\begin{equation}
\label{eq:lift_of_AimpliesB}
L(A \rightarrow B) = \frac{C(A \rightarrow B)}{S(B)} = \frac{P(A \cap B)}{P(A) \cdot P(B)}
\end{equation}\\

To ensure the robustness of the discovered rules, we set the minimum support, confidence, and lift based on the achieved cluster number and data size proportion. The ARM algorithm applied in our data is expressed through Algorithm \ref{alg:arm_algorithm}.

\begin{algorithm}[H]
\caption{Cluster-based ARM}
\label{alg:arm_algorithm}
\begin{algorithmic}[1]
\Require Encoded dataset $N$
\Require Cluster labels $\{0,1,\ldots,K-1\}$, where $K$ is the number of clusters (K-means)
\Require Thresholds $Th = [s_{\min}, c_{\min}, L_{\min}, RL_{\max}]$
\Ensure Rules per cluster

\For{$k \gets 0$ \textbf{to} $K-1$}
    \State Extract subset $N_k \subset N$ containing records in cluster $k$
    \State Filter out sparse columns in $N_k$
    \State Convert $N_k$ to boolean format (if needed)
    \State Apply Apriori on $N_k$ to obtain frequent itemsets with support $\ge s_{\min}$
    \State Generate rules from frequent itemsets subject to $c_{\min}$, $L_{\min}$, and $RL_{\max}$
    \ForAll{candidate rules $X$ with $\mathrm{Len}(X)\le RL_{\max}$}
        \State Compute support $S(X)$, confidence $C(X)$, and lift $L(X)$
        \If{$S(X)\ge s_{\min}$ \textbf{and} $C(X)\ge c_{\min}$ \textbf{and} $L(X)\ge L_{\min}$}
            \State Store rule $X$ for cluster $k$
        \EndIf
    \EndFor
\EndFor
\end{algorithmic}
\end{algorithm}



\section{RESULTS}
\subsection{K-means Clustering}
\begin{figure}[!ht]
  \centering
  \includegraphics[width=0.6\textwidth, height=0.28\textheight]{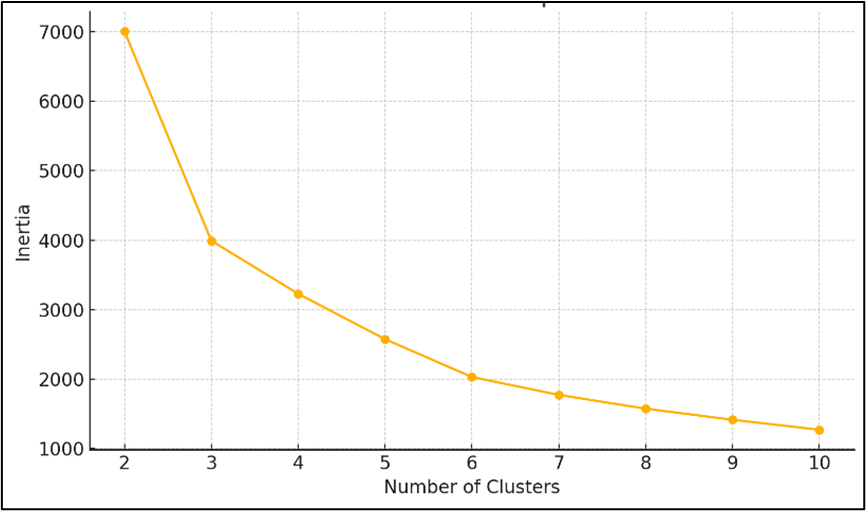}
  \caption{Optimum Cluster Identification}
  \label{fig:optimum_cluster_identification}
\end{figure}

\begin{figure}[!ht]
  \centering
  \includegraphics[width=0.9\textwidth, height=0.9\textheight]{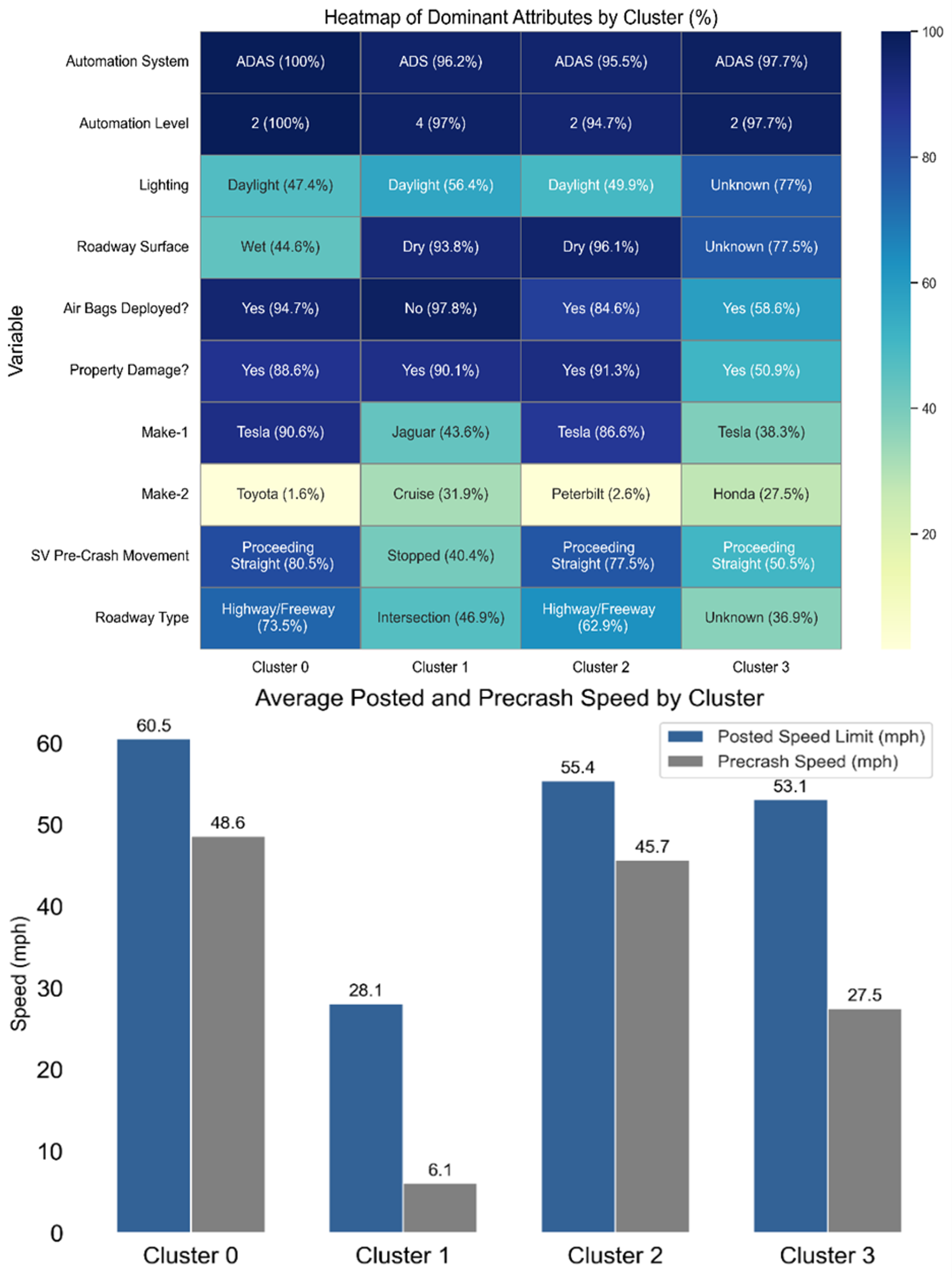}
  \caption{Cluster Profile}
  \label{fig:clusterprofile}
\end{figure}

We use the elbow method to identify the optimum number of clusters. The elbow method is a popular method for clustering approaches like K-means, which involves plotting WCSS against cluster numbers and identifying an elbow point beyond which adding more clusters does not benefit the analysis \cite{bholowalia2014ebk}. From Figure \ref{fig:optimum_cluster_identification}, it is evident that the optimum cluster number is K = 4. The cluster characteristics of these 4 clusters are explained in detail in Figure \ref{fig:clusterprofile}. Cluster 1 is the only cluster that is ADS dominant (SAE Level 4), and all other clusters are ADAS system influenced with SAE Level 2. It is evident that 40\% of the ADS-equipped Level 4 vehicles stopped before crashes, which might be issues with their road or surrounding recognition system. These crashes occurred mostly at intersections (47\%). Clusters 0, 2, and 3 are all ADAS-dominated SAE Level 2 Teslas operating on a high-speed highway and crashed while proceeding straight.  


\subsection{Association Rule Mining(ARM)}
While clustering provides us with insightful information about the AV crash pattern, it usually includes factors that go together. However, how those factors are related to each other, this multivariate relationship, can be investigated by generating associations/rules through ARM on each cluster. We use Apriori with the following thresholds Support $\ge$ 0.05, Confidence $\ge$ 0.6, Lift $\ge$ 1.2, and Maximum rule length $\le$ 3.
	
A support value of 0.05 indicates an association rule will apply to a minimum of 5\% of the rows in the subset of the cluster to be considered. Our final dataset has a size of $2380 \times 31$ and 4 clusters are established, indicating 595 records for each cluster. 5\% support leads us to the rule or pattern appearing in at least 26-30 records per cluster, which provides us a good opportunity to capture a meaningful pattern. A confidence above 0.5 proves the rule is more likely to be true than false, which is why 0.6 is a safe choice, ensuring the generated rules will be true for at least 60\% of the time. A lift of 1.2 is also a promising choice, as any lift value greater than 1 shows a strong positive association. For computational efficiency and to avoid overfitting, we use 3-item rules and limit maximum rule length to 3. We apply these thresholds and generate 9165 different rules (Cluster 0: 2753, Cluster 1: 1472, Cluster 2: 719, and Cluster 3: 4221). These rules are then ranked according to their strength of associations (lift), and the top 10 rules for each cluster were summarized for interpretation.\\

Table \ref{tab:cluster_0_rules} shows ADAS-dominated Cluster 0 rules where crash patterns are strongly influenced by weather and road surface conditions. With 71\% confidence and a lift of 3.93, rule 3 denotes that clear weather is frequently and strongly associated with crashes on dry roadways, particularly when the AV is proceeding straight on highways. Rule 4 says multi-front impacts tend to occur 81\% of the time in cloudy and rainy weather, even when roads are otherwise described as having no unusual conditions. Rule 7 shows new vehicles with low mileage tend to be involved in crashes when the weather is cloudy and rainy. Rule 9 indicates that most of the crashes occurred in March and hit a fixed object, although the road condition was usual and there was no missing marking or work zone. Fixed-object collisions occur more frequently when crashes involve wet surfaces and front multi-zone impacts and are also observed in March under normal road conditions. Moreover, multi-front impacts are linked to airbag deployment and inclement weather, indicating that during inclement weather situations, crashes involving front impact often deploy airbags. Nighttime crashes on well-lit roads are typically associated with wet surfaces.

\begin{table}[H]
\centering
\caption{Association Rules for Cluster 0 (SAE Level 2 with ADAS)}
\label{tab:cluster_0_rules}
\begin{adjustbox}{width=\textwidth}
\begin{tabular}{|c|p{10cm}|c|c|c|}
\hline
\textbf{Sl.} & \textbf{Rule} & \textbf{Support} & \textbf{Confidence} & \textbf{Lift} \\
\hline
1 & [Weather = Clear] $\rightarrow$ [Roadway Surface = Dry] + [SV Pre-Crash Movement = Proceeding Straight] & 0.06 & 0.63 & 4.37 \\
\hline
2 & [Roadway Type = Highway / Freeway] + [Weather = Clear] $\rightarrow$ [Roadway Surface = Dry] & 0.05 & 0.71 & 3.93 \\
\hline
3 & [SV Contact Area = Front Left, Front \& Front Right] + [Roadway Description = No Unusual Conditions] $\rightarrow$ [Weather = Cloudy \& Rain] & 0.10 & 0.81 & 2.71 \\
\hline
4 & [Weather = Cloudy \& Rain] $\rightarrow$ [Roadway Surface = Wet] + [Roadway Description = No Unusual Conditions] & 0.27 & 0.90 & 2.36 \\
\hline
5 & [Reporting Entity = Tesla, Inc.] + [Weather = Cloudy \& Rain] $\rightarrow$ [Roadway Surface = Wet] & 0.30 & 0.99 & 2.22 \\
\hline
6 & [SV Contact Area = Front Left, Front \& Front Right] + [Roadway Surface = Wet] $\rightarrow$ [Crash With = Other Fixed Object] & 0.11 & 0.81 & 2.15 \\
\hline
7 & [Mileage Bin = 30k–60k] + [Roadway Description = No Unusual Conditions] $\rightarrow$ [Weather = Cloudy \& Rain] & 0.12 & 0.64 & 2.14 \\
\hline
8 & [SV Contact Area = Front Left, Front \& Front Right] $\rightarrow$ [SV Any Air Bags Deployed? = Yes] + [Weather = Cloudy \& Rain] & 0.18 & 0.61 & 2.08 \\
\hline
9 & [Month = March] + [Crash With = Other Fixed Object] $\rightarrow$ [Roadway Description = No Unusual Conditions] & 0.05 & 0.90 & 2.08 \\
\hline
10 & [Lighting = Dark - Lighted] + [Roadway Description = No Unusual Conditions] $\rightarrow$ [Roadway Surface = Wet] & 0.06 & 0.92 & 2.08 \\
\hline
\end{tabular}
\end{adjustbox}
\end{table}

Table \ref{tab:cluster_1_rules} represents Cluster 1. Here, crash patterns are dominated by rear-end collisions involving vehicles equipped with ADS, high-level automation. While cluster 0 ADAS-involved crashes resulted in frontal damage of subject AV, ADS-involved crashes in cluster 1 mostly revolve around rear side damage of subject AV and frontal damage of the crash partner vehicles. These incidents often occur on dry roads under otherwise normal driving conditions, with a notable concentration of crashes reported by Zoox involving Toyota vehicles equipped with Automated Driving Systems (ADS)(Rule 2, 3). Additionally, many of these crashes occur while the contact vehicle is proceeding straight, and surprisingly, they often do not deploy the airbag in the subject vehicle. A substantial share of the contact vehicles are associated with SAE Level 4 automation(Rule 9), indicating that this cluster likely reflects routine operational failures during rear-end encounters in mixed traffic conditions.

\begin{table}[H]
\centering
\caption{Association Rules for Cluster 1 (SAE Level 4 with ADS)}
\label{tab:cluster_1_rules}
\begin{adjustbox}{width=\textwidth}
\begin{tabular}{|c|p{10cm}|c|c|c|}
\hline
\textbf{Sl.} & \textbf{Rule} & \textbf{Support} & \textbf{Confidence} & \textbf{Lift} \\
\hline
1 & [Roadway Surface = Dry] + [Reporting Entity = Zoox, Inc.] $\rightarrow$ [Make = Toyota] & 0.08 & 1.00 & 13.26 \\
\hline
2 & [Reporting Entity = Zoox, Inc.] + [Roadway Description = No Unusual Conditions] $\rightarrow$ [Make = Toyota] & 0.06 & 1.00 & 13.26 \\
\hline
3 & [Make = Toyota] $\rightarrow$ [Automation System = ADS] + [Reporting Entity = Zoox, Inc.] & 0.05 & 0.77 & 13.26 \\
\hline
4 & [SV Contact Area = Rear Left, Rear \& Rear Right] + [Make = Cruise] $\rightarrow$ [Roadway Type = Intersection] & 0.05 & 0.94 & 9.20 \\
\hline
5 & [SV Contact Area = Rear Left, Rear \& Rear Right] $\rightarrow$ [CP Contact Area = Front Left \& Front \& Front Right] + [Make = Jaguar] & 0.05 & 0.61 & 8.81 \\
\hline
69 & [CP Contact Area = Front] $\rightarrow$ [Roadway Surface = Dry] + [SV Contact Area = Rear] & 0.07 & 0.65 & 8.17 \\
\hline
7 & [SV Contact Area = Rear \& Rear Right] $\rightarrow$ [Lighting = Daylight] + [CP Pre-Crash Movement = Proceeding Straight] & 0.07 & 0.79 & 8.38 \\
\hline
8 & [CP Contact Area = Front Left, Front \& Front Right] $\rightarrow$ [SV Any Air Bags Deployed? = No] + [Roadway Type = Intersection] & 0.06 & 0.66 & 8.08 \\
\hline
9 & [CP Contact Area = Front Left and Front] $\rightarrow$ [Automation Level = 4] + [SV Contact Area = Rear] & 0.05 & 0.67 & 8.10 \\
\hline
10 & [CP Contact Area = Front] $\rightarrow$ [SV Contact Area = Rear] & 0.05 & 0.68 & 8.13 \\
\hline
\end{tabular}
\end{adjustbox}
\end{table}

From the cluster profile in Figure \ref{fig:clusterprofile}, we know Cluster 2 is mostly ADAS-equipped Level 2 vehicles operating on highway/freeway. In Cluster 2 (shown in Table \ref{tab:cluster_2_rules}), crash patterns are centered around rear-end collisions occurring mainly on highways that result in little to no injury. A frequent scenario involves the contact vehicle striking the rear of another vehicle, often resulting in front-end damage to the subject vehicle, especially in cases reported by Tesla. These crashes frequently occur under normal road conditions and on freeway segments. Notably, many of the involved vehicles are Teslas, and several rules suggest strong directional patterns, such as rules 9 and 10, where Tesla vehicles with front-end damage are often linked to rear-end damage on the opposing vehicle. Property damage is common, but injury severity is generally low, and airbag deployment is frequently absent, suggesting lower-impact or controlled-speed collisions.

\begin{table}[H]
\centering
\caption{Association Rules for Cluster 2 (SAE Level 2 with ADAS)}
\label{tab:cluster_2_rules}
\begin{adjustbox}{width=\textwidth}
\begin{tabular}{|c|p{10.5cm}|c|c|c|}
\hline
\textbf{Sl.} & \textbf{Rule} & \textbf{Support} & \textbf{Confidence} & \textbf{Lift} \\
\hline
1 & [Highest Injury Severity Alleged = No Injuries Reported] $\rightarrow$ [SV Any Air Bags Deployed? = No] + [SV Pre-Crash Movement = Proceeding Straight] & 0.06 & 0.62 & 5.71 \\
\hline
2 & [Roadway Type = Highway / Freeway] + [Highest Injury Severity Alleged = No Injuries Reported] $\rightarrow$ [SV Any Air Bags Deployed? = No] & 0.05 & 0.77 & 5.55 \\
\hline
3 & [CP Contact Area = Rear Left, Rear \& Rear Right] $\rightarrow$ [SV Contact Area = Front Left, Front \& Front Right] + [Roadway Type = Highway / Freeway] & 0.06 & 0.75 & 5.37 \\
\hline
4 & [Highest Injury Severity Alleged = No Injuries Reported] + [Property Damage? = Yes] $\rightarrow$ [SV Any Air Bags Deployed? = No] & 0.06 & 0.74 & 5.33 \\
\hline
5 & [Highest Injury Severity Alleged = No Injuries Reported] $\rightarrow$ [SV Any Air Bags Deployed? = No] + [Roadway Description = No Unusual Conditions] & 0.05 & 0.65 & 5.27 \\
\hline
6 & [Roadway Type = Highway / Freeway] + [SV Contact Area = Rear Left, Left \& Front Left] $\rightarrow$ [Crash With = Other Fixed Object] & 0.05 & 0.87 & 4.51 \\
\hline
7 & [SV Contact Area = Rear Left, Left \& Front Left] + [Property Damage? = Yes] $\rightarrow$ [Crash With = Other Fixed Object] & 0.05 & 0.80 & 4.15 \\
\hline
8 & [SV Contact Area = Rear Left] $\rightarrow$ [Automation Level = 2] + [Crash With = Other Fixed Object] & 0.05 & 0.81 & 4.22 \\
\hline
9 & [Make = Tesla] + [CP Contact Area = Rear Left, Rear \& Rear Right] $\rightarrow$ [SV Contact Area = Front Left, Front \& Front Right] & 0.07 & 0.95 & 3.44 \\
\hline
10 & [CP Contact Area = Rear Left, Rear \& Rear Right] $\rightarrow$ [SV Contact Area = Front Left, Front \& Front Right] + [Reporting Entity = Tesla, Inc.] & 0.07 & 0.86 & 3.32 \\
\hline
\end{tabular}
\end{adjustbox}
\end{table}

Cluster 3 is shown in Table \ref{tab:cluster_3_rules}. This is also an ADAS-dominated cluster where crashes usually occur under standard driving conditions and are strongly associated with Level 2 automated vehicles, telematics-based reporting, and major manufacturers such as Honda and Tesla. Rules 2 and 3 have strong lift values of 4.49 and 4.4, respectively, indicating most of the accidents that were recorded for Level 2 ADAS AV vehicles occurred during daylight. From rule 8 with a strong lift of 3.08, we notice that most of the Honda-reported crashes when they got hit on the front resulted in airbag deployment. Rule 10 also shows that most of the Honda-reported crashes with SAE Level 2 automation result in moderate-level injury severity. Tesla reports, on the other hand, are predominantly submitted via telematics and frequently lack detailed injury documentation. Vehicles in the 30,000–60,000 mileage range, when airbags deploy, also tend to be reported through telematics, suggesting reliance on sensor-triggered reporting in maturing AV fleets. Several rules highlight the involvement of ADAS-equipped or Level 2 AVs operating in daylight on otherwise normal roads, yet these vehicles still encounter crashes on wet surfaces.

\begin{table}[H]
\centering
\caption{Association Rules for Cluster 3 (SAE Level 2 with ADAS)}
\label{tab:cluster_3_rules}
\begin{adjustbox}{width=\textwidth}
\begin{tabular}{|c|p{10.5cm}|c|c|c|}
\hline
\textbf{Sl.} & \textbf{Rule} & \textbf{Support} & \textbf{Confidence} & \textbf{Lift} \\
\hline
1 & [Automation System = ADAS] + [Roadway Description = No Unusual Conditions] $\rightarrow$ [Roadway Surface = Wet] & 0.06 & 0.68 & 6.60 \\
\hline
2 & [Automation Level = 2] + [Roadway Description = No Unusual Conditions] $\rightarrow$ [Lighting = Daylight] & 0.07 & 0.79 & 4.49 \\
\hline
3 & [Roadway Description = No Unusual Conditions] $\rightarrow$ [Air Bags Deployed? = Yes] & 0.08 & 0.77 & 4.40 \\
\hline
4 & [Roadway Description = No Unusual Conditions] $\rightarrow$ [Lighting = Daylight] + [Automation System = ADAS] & 0.07 & 0.68 & 4.20 \\
\hline
5 & [Roadway Description = No Unusual Conditions] $\rightarrow$ [Lighting = Daylight] + [Automation Level = 2] & 0.07 & 0.68 & 4.20 \\
\hline
6 & [Source = Telematics] $\rightarrow$ [Highest Injury Severity Alleged = No Serious Injury] + [Reporting Entity = Tesla, Inc.] & 0.24 & 0.96 & 3.24 \\
\hline
7 & [Mileage Bin = 30k--60k] + [SV Any Air Bags Deployed? = Yes] $\rightarrow$ [Source = Telematics] & 0.08 & 0.78 & 3.10 \\
\hline
8 & [SV Contact Area = Front] $\rightarrow$ [SV Any Air Bags Deployed? = Yes] + [Reporting Entity = Honda (American Honda Motor Co.)] & 0.05 & 0.67 & 3.08 \\
\hline
9 & [Source = Telematics] $\rightarrow$ [SV Any Air Bags Deployed? = Yes] + [Reporting Entity = Tesla, Inc.] & 0.22 & 0.88 & 3.08 \\
\hline
10 & [Highest Injury Severity Alleged = Moderate] $\rightarrow$ [Automation Level = 2] + [Reporting Entity = Honda (American Honda Motor Co.)] & 0.08 & 0.90 & 2.56 \\
\hline
\end{tabular}
\end{adjustbox}
\end{table}

\section{DISCUSSION}
We applied an unsupervised machine learning clustering method, K-means, and Association Rule Mining to reveal hidden patterns of AV crashes with different levels of automation and different driving assistance systems such as ADS \& ADAS After processing the raw data, we feed it to the hybrid model framework. First, we cluster the dataset and optimize the cluster number to 4. Cluster 0, Cluster 2, and Cluster 3 are mostly dominated by automation Level 2 AV-involved crashes equipped with ADAS, and Cluster 1 is dominated by Level 4 AV crashes equipped with ADS. The cluster profile initially reveals very interesting insights. It proves that in ADAS-involved crashes (especially Cluster 0 and Cluster 2), AV vehicles mostly operated on the highways or freeways, and in most cases, airbags were deployed and resulted in property damage (almost 90\% of the time). This shows there could still be the human element responsible for the crashes, as drivers have the primary responsibility and ADAS was not enough to prevent crashes. Most of these AVs are Tesla (85-90\%) and had an average pre-crash speed of 45 mph while the speed limit was 55 mph to 65 mph. This shows that even though they resulted in crashes, they were not speeding, which explains the higher percentage of non-injury to moderate-injury crashes. Cluster 3 records show a good percentage of unknown factors, indicating the lack of data or reporting irregularities. Cluster 1 is representative of high-automation Level 4 (with ADS) AV crashes. We noticed the average posted speed limit of these is 28 mph, which indicates most of these crashes occurred in urban streets or in a controlled environment. Almost 50\% of the time, crashes occur at intersections, showing Level 4 still struggles in mixed traffic with higher conflict points. Cluster 1 also shows these Level 4 AVs (with ADS) came at a very slow speed before the crash and did not deploy airbags. This makes sense, as most of these crashes did not cause injury.\\

We applied ARM on each of these clusters and found exciting information on how wide ranges of crash factors are related to each other and the strength of those relationships. We narrowed down the top 10 rules based on the lift for all 4 clusters. Cluster 0 shows the contribution of environmental factors to Level 2 AV (with ADAS) crashes. For example, during rainy conditions, subject AVs usually hit fixed objects and damage the front, front-left, and front-right sides of the car. Rules also show that when crashes on this cluster occur during the night, the roads are wet even though the street is properly lit. It indicates that the primary reason may be due to the inability of the driver to drive properly on wet roads, as in ADAS equipped vehicles, drivers still hold primary control. Cluster 1 rules identify different relationships of Level 4 AV (with ADS) related crashes involving important vehicle dynamics like pre-crash movements of AV, ADS system awareness, and damage pattern. Rules indicate that in most crashes when subject AVs had damage on the rear side, the crash partner vehicle had frontal damage, and most of the time, the car was a jaguar. This indicates that in mixed traffic conditions of urban streets, ADS systems in Jaguars are prone to rear-end collisions. Rules also express that during crashes of Level 2 AVs, when crash partner vehicles were hit on the front, airbags did not deploy in the subject vehicle, and this scenario is common at urban intersections. Rules in Cluster 2 express the directional trend of ADAS-equipped Level 2 Tesla AV crashes, where if AVs take front-end damages, the crash partner vehicles end up with rear-end damages. Cluster 3 rules reflect more routine driving situations rather than complex settings. Rules indicate that when accidents occur in normal road conditions, it is often daylight and airbags are deployed. We also notice that when the mileage of the AV vehicles was between 30k and 60k and airbags were deployed, the accident had been reported via a telematics source. Overall, the generated rules provide interesting interrelations of environmental factors, road conditions, lighting, automation systems, mileage, vehicle maneuvers, etc., which enable context-specific decision-making.

\section{CONCLUSIONS}
In this paper, we propose a hybrid framework of clustering and rule-mining to analyze multivariate relationships of crash factors in SAE Level 2 (with ADAS) and SAE Level 4 (with ADS) AV crashes. Our architecture uses K-means to cluster large-scale AV crash data into 4 different clusters of similar trends. We mine these clusters with the Apriori algorithm to generate different rules and find multivariate dependencies across a wide range of crash factors. We select top associations from each cluster with the strongest likelihood of occurrence. Results reveal that ADS-equipped Level 4 AVs are usually involved in crashes on urban streets, especially at intersections, showing their limited capacity to capture complex traffic behavior in mixed traffic. These AVs show a rear-end collision trend, while their crash partner vehicles take front-side collision damage. Level 2 AVs with ADAS systems mostly operate on highways and tend to be involved in crashes during bad weather and wet road conditions, mostly taking front damage. Extracted ARM rules with strong likelihood offer us a granular understanding of automated vehicle crash patterns with mixed automation levels and driving assistance systems. Consideration of these intricate dependencies of vehicle dynamics, environmental factors, road conditions, and automation systems will benefit traffic engineers, AV developers, researchers, and policy makers in identifying high-risk factors, validating AV automation systems, planning smarter ITS (Intelligent Transportation System) infrastructure, and developing evidence-based local, state, or federal policy and safety standards.\\

The limitation of our study is the high percentage of unknown information within few variables and some confidential crash narratives, which were not analyzed on a granular scale to extract insights. Future research will include these narratives for text mining to fill up the information gap among other variables. We will also include other crash data sources in conjunction with NHTSA to enhance scalability and robustness.

\section{AUTHOR CONTRIBUTIONS}
The authors confirm contribution to the paper as follows: study conception and design: J.R. Palit, and O.A. Osman; data collection: J.R. Palit; analysis and interpretation of results: J.R. Palit, V.K. Kumarasamy, and O.A. Osman; draft manuscript preparation: J.R. Palit, V.K. Kumarasamy, and O.A. Osman. All authors reviewed the results and approved the final version of the manuscript.
\newpage

\bibliographystyle{trb}

\begin{thebibliography}{32}
\providecommand{\natexlab}[1]{#1}

\bibitem[{of~Transportation(2024)}]{USDOT_2024_new}
of~Transportation, U.~D., \emph{The Roadway Safety Problem}. U.S. Department of Transportation, 2024, accessed July 20, 2025.

\bibitem[{Administration(2025)}]{NHTSA_2025_new}
Administration, N. H. T.~S., \emph{Automated Vehicles for Safety}. National Highway Traffic Safety Administration, 2025, accessed July 20, 2025.

\bibitem[{International(2025)}]{Int_S_Taxonomy_2025_new}
International, S., \emph{Taxonomy and Definitions for Terms Related to Driving Automation Systems for On-Road Motor Vehicles}. SAE International, 2025, sAE Standard J3016.

\bibitem[{Lee et~al.(2020)Lee, Hwang, Kang, and Song}]{lee2020black}
Lee, H., K.~Hwang, M.~Kang, and J.~Song, Black ice detection using CNN for the Prevention of Accidents in Automated Vehicle. In \emph{2020 International Conference on Computational Science and Computational Intelligence (CSCI)}, IEEE, 2020, pp. 1189--1192.

\bibitem[{Lee et~al.(2024)Lee, Kang, Hwang, and Yoon}]{lee2024typical}
Lee, H., M.~Kang, K.~Hwang, and Y.~Yoon, The typical AV accident scenarios in the urban area obtained by clustering and association rule mining of real-world accident reports. \emph{Heliyon}, Vol.~10, No.~3, 2024.

\bibitem[{Schoettle and Sivak(2015)}]{schoettle2015preliminary}
Schoettle, B. and M.~Sivak, A preliminary analysis of real-world crashes involving self-driving vehicles. \emph{University of Michigan Transportation Research Institute}, 2015.

\bibitem[{Favar{\`o} et~al.(2017)Favar{\`o}, Nader, Eurich, Tripp, and Varadaraju}]{favaro2017examining}
Favar{\`o}, F.~M., N.~Nader, S.~O. Eurich, M.~Tripp, and N.~Varadaraju, Examining accident reports involving autonomous vehicles in California. \emph{PLoS one}, Vol.~12, No.~9, 2017, p. e0184952.

\bibitem[{Alambeigi et~al.(2020)Alambeigi, McDonald, and Tankasala}]{alambeigi2020crash}
Alambeigi, H., A.~D. McDonald, and S.~R. Tankasala, Crash themes in automated vehicles: A topic modeling analysis of the California Department of Motor Vehicles automated vehicle crash database. \emph{arXiv preprint arXiv:2001.11087}, 2020.

\bibitem[{Ren et~al.(2021)Ren, Yu, Chen, Gao, and Bao}]{ren2021divergent}
Ren, W., B.~Yu, Y.~Chen, K.~Gao, and S.~Bao, Divergent effects of factors on crashes under autonomous and conventional driving modes using a hierarchical bayesian approach. \emph{arXiv preprint arXiv:2108.02422}, 2021.

\bibitem[{Boggs et~al.(2020)Boggs, Wali, and Khattak}]{boggs2020exploratory}
Boggs, A.~M., B.~Wali, and A.~J. Khattak, Exploratory analysis of automated vehicle crashes in California: A text analytics \& hierarchical Bayesian heterogeneity-based approach. \emph{Accident Analysis \& Prevention}, Vol. 135, 2020, p. 105354.

\bibitem[{Kutela et~al.(2022)Kutela, Das, and Dadashova}]{kutela2022mining}
Kutela, B., S.~Das, and B.~Dadashova, Mining patterns of autonomous vehicle crashes involving vulnerable road users to understand the associated factors. \emph{Accident Analysis \& Prevention}, Vol. 165, 2022, p. 106473.

\bibitem[{Lee et~al.(2023)Lee, Arvin, and Khattak}]{lee2023advancing}
Lee, S., R.~Arvin, and A.~J. Khattak, Advancing investigation of automated vehicle crashes using text analytics of crash narratives and Bayesian analysis. \emph{Accident Analysis \& Prevention}, Vol. 181, 2023, p. 106932.

\bibitem[{Dunson(2009)}]{dunson2009bayesian}
Dunson, D.~B., Bayesian nonparametric hierarchical modeling. \emph{Biometrical Journal: Journal of Mathematical Methods in Biosciences}, Vol.~51, No.~2, 2009, pp. 273--284.

\bibitem[{Das et~al.(2020)Das, Dutta, and Tsapakis}]{das2020automated}
Das, S., A.~Dutta, and I.~Tsapakis, Automated vehicle collisions in California: Applying Bayesian latent class model. \emph{IATSS research}, Vol.~44, No.~4, 2020, pp. 300--308.

\bibitem[{Kang et~al.(2020)Kang, Hyndman, and Li}]{kang2020gratis}
Kang, Y., R.~J. Hyndman, and F.~Li, GRATIS: GeneRAting TIme Series with diverse and controllable characteristics. \emph{Statistical Analysis and Data Mining: The ASA Data Science Journal}, Vol.~13, No.~4, 2020, pp. 354--376.

\bibitem[{Kumbhare and Chobe(2014)}]{kumbhare2014overview}
Kumbhare, T.~A. and S.~V. Chobe, An overview of association rule mining algorithms. \emph{International Journal of Computer Science and Information Technologies}, Vol.~5, No.~1, 2014, pp. 927--930.

\bibitem[{Qin and Liu(2025)}]{qin2025cracking}
Qin, S. and L.~Liu, Cracking the Code of Car Crashes: How Autonomous and Human Driving Differ in Risk Factors. \emph{Sustainability}, Vol.~17, No.~10, 2025, p. 4368.

\bibitem[{Ashraf et~al.(2021)Ashraf, Dey, Mishra, and Rahman}]{ashraf2021extracting}
Ashraf, M.~T., K.~Dey, S.~Mishra, and M.~T. Rahman, Extracting rules from autonomous-vehicle-involved crashes by applying decision tree and association rule methods. \emph{Transportation research record}, Vol. 2675, No.~11, 2021, pp. 522--533.

\bibitem[{Li et~al.(2025)Li, Wang, Wang, and Liu}]{li2025characteristics}
Li, Y., X.~Wang, T.~Wang, and Q.~Liu, Characteristics analysis of autonomous vehicle pre-crash scenarios. \emph{arXiv preprint arXiv:2502.20789}, 2025.

\bibitem[{Ding et~al.(2024)Ding, Abdel-Aty, Barbour, Wang, Wang, and Zheng}]{ding2024exploratory}
Ding, S., M.~Abdel-Aty, N.~Barbour, D.~Wang, Z.~Wang, and O.~Zheng, Exploratory analysis of injury severity under different levels of driving automation (SAE Levels 2 and 4) using multi-source data. \emph{Accident Analysis \& Prevention}, Vol. 206, 2024, p. 107692.

\bibitem[{Yan et~al.(2024)Yan, Huang, and He}]{yan2024comparison}
Yan, S., C.~Huang, and D.~He, A comparison of patterns and contributing factors of ADAS and ADS involved crashes. \emph{Journal of Transportation Safety \& Security}, Vol.~16, No.~9, 2024, pp. 1061--1088.

\bibitem[{GOV(2025)}]{NHTSA_gov_2025}
GOV, N., \emph{NHTSA. Standing General Order on Crash Reporting. [cited 2024 7/20/2024]}. NHTSA GOV, 2025.

\bibitem[{Kaiser(2014)}]{kaiser2014dealing}
Kaiser, J., Dealing with Missing Values in Data. \emph{Journal Of Systems Integration (1804-2724)}, Vol.~5, No.~1, 2014.

\bibitem[{Kim and Curry(1977)}]{kim1977treatment}
Kim, J.-O. and J.~Curry, The treatment of missing data in multivariate analysis. \emph{Sociological Methods \& Research}, Vol.~6, No.~2, 1977, pp. 215--240.

\bibitem[{Jain(2010)}]{jain2010data}
Jain, A.~K., Data clustering: 50 years beyond K-means. \emph{Pattern recognition letters}, Vol.~31, No.~8, 2010, pp. 651--666.

\bibitem[{Xu and Wunsch(2005)}]{xu2005survey}
Xu, R. and D.~Wunsch, Survey of clustering algorithms. \emph{IEEE Transactions on neural networks}, Vol.~16, No.~3, 2005, pp. 645--678.

\bibitem[{Morissette et~al.(2013)Morissette, Chartier et~al.}]{morissette2013k}
Morissette, L., S.~Chartier, et~al., The k-means clustering technique: General considerations and implementation in Mathematica. \emph{Tutorials in Quantitative Methods for Psychology}, Vol.~9, No.~1, 2013, pp. 15--24.

\bibitem[{Burkardt(2009)}]{burkardt2009k}
Burkardt, J., K-means clustering. \emph{Virginia Tech, Advanced Research Computing, Interdisciplinary Center for Applied Mathematics}, Vol.~5, 2009.

\bibitem[{Cui et~al.(2020)}]{cui2020introduction}
Cui, M. et~al., Introduction to the k-means clustering algorithm based on the elbow method. \emph{Accounting, Auditing and Finance}, Vol.~1, No.~1, 2020, pp. 5--8.

\bibitem[{Hahsler et~al.(2015)Hahsler, Gruen, Hornik, and Buchta}]{hahsler2015mining}
Hahsler, M., B.~Gruen, K.~Hornik, and C.~Buchta, \emph{Mining association rules and frequent itemsets. R package version 1.3-1}, 2015.

\bibitem[{Hahsler(2019)}]{hahsler2019arulesviz}
Hahsler, M., \emph{arulesViz: Visualizing Association Rules and Frequent Itemsets. R package version 1.3-3}, 2019.

\bibitem[{Bholowalia and Kumar(2014)}]{bholowalia2014ebk}
Bholowalia, P. and A.~Kumar, EBK-means: A clustering technique based on elbow method and k-means in WSN. \emph{International Journal of Computer Applications}, Vol. 105, No.~9, 2014.

\end{thebibliography}

\end{document}